\documentclass[conference,a4]{IEEEtran}
\usepackage[utf8]{inputenc}
\usepackage[T1]{fontenc}
\usepackage{graphicx}
\usepackage{float}
\usepackage[font=footnotesize,labelfont=bf]{caption}  
\usepackage{amsmath, amssymb}
\usepackage{hyperref}
\hypersetup{colorlinks=true, linkcolor=black, citecolor=black, urlcolor=black}
\usepackage{booktabs}
\usepackage{tabularx}
\usepackage{adjustbox}
\usepackage[nottoc]{tocbibind}
\usepackage{xurl}  
\usepackage{multirow}
\usepackage{makecell}

\usepackage{subcaption} 

\usepackage{siunitx}

\usepackage{fancyhdr}

\usepackage{orcidlink}

\begin{document}

\makeatletter
\def\ps@headings{%
  \let\@oddhead\@empty
  \let\@evenhead\@empty
  \let\@oddfoot\@empty
  \let\@evenfoot\@empty
}
\def\ps@IEEEtitlepagestyle{%
  \let\@oddhead\@empty
  \let\@evenhead\@empty
  \let\@oddfoot\@empty
  \let\@evenfoot\@empty
}
\makeatother

\title{A Novel CNN–Gradient Boosting Ensemble for Guava Disease Detection}

\author{
    \IEEEauthorblockN{Tamim Ahasan Rijon}
    \IEEEauthorblockA{
        Department of Computer Science and Engineering \\
        Daffodil International University \\
        Ashulia, Dhaka 1341, Bangladesh \\
        Email: tamimahasan.ar@gmail.com
    }
    \and
    \IEEEauthorblockN{Yeasin Arafath\orcidlink{0009-0001-2355-1046}}
    \IEEEauthorblockA{
        Department of Computer Science and Engineering \\
        Brac University \\
        Merul Badda, Dhaka 1212, Bangladesh \\
        Email: yeasin.cse.007@gmail.com
    }
}

\maketitle

\begin{abstract}

As a significant agricultural country, Bangladesh utilizes its fertile land for guava cultivation and dedicated labor to boost its economic development. In a nation like Bangladesh, enhancing guava production and agricultural practices plays a crucial role in its economy. Anthracnose and fruit fly infection can lower the quality and productivity of guava, a crucial tropical fruit. Expert systems that detect diseases early can reduce losses and safeguard the harvest. Images of guava fruits classified into the Healthy, Fruit Flies, and Anthracnose classes are included in the Guava Fruit Disease Dataset 2024 (GFDD24), which comes from plantations in Rajshahi and Pabna, Bangladesh. This study aims to create models using CNN alongside traditional machine learning techniques that can effectively identify guava diseases in locally cultivated varieties in Bangladesh. In order to achieve the highest classification accuracy of $\sim$99.99\% for the guava dataset, we propose utilizing ensemble models that combine CNN-ML with Gradient Boosting Machine. In general, the CNN–ML cascade framework exhibits strong, high-accuracy guava disease detection that is appropriate for real-time agricultural monitoring systems.

\end{abstract}


\section{INTRODUCTION}

Guava cultivation plays a significant role in Bangladeshi agriculture as it provides both food and a source of income for many people \cite{r10}. However, guava trees are susceptible to various diseases that can severely impact both the yield and quality of the fruit \cite{r1}. Timely and precise detection of diseases is crucial for implementing effective strategies to minimize losses and prevent outbreaks. Traditional visual inspection techniques often rely on subjective judgment and prove to be inadequate, particularly for new plantations \cite{r10}.

In Asian countries like Pakistan, guava is a vital crop that encounters various organic and fungal diseases, leading to considerable reductions in output. Accurate analysis of diseases is necessary due to the subtle differences in symptoms, and misdiagnosis can lead to financial repercussions for farmers. Thus, there is a requirement for computer-vision-based monitoring of guava plants in the field to aid in identifying and classifying diseases \cite{r1}. Guava, a widely consumed fruit in numerous countries, ranks as the fourth most important fruit in Pakistan. Nonetheless, the presence of bacterial and fungal diseases significantly hinders guava production, causing ongoing declines in yield.

Guava is especially vulnerable to a variety of fungal, bacterial, and postharvest infections. Plant diseases are a serious danger to agricultural output, lowering both the quality and quantity of crops.  To avoid large financial losses brought on by incorrect diagnoses or postponed treatment, accurate and prompt disease identification is crucial.  Automated monitoring technologies that can accurately identify and categorize guava infections are desperately needed to assist efficient management and sustain high crop output.  By creating an automated framework for accurate and timely guava disease diagnosis, our research helps to meet this requirement.

The research employed conventional machine learning techniques alongside CNN feature extraction to identify diseases in guavas. EfficientNet-B0 was utilized for extracting features from images, which were subsequently trained through ML Gradient Boosting. Ensemble cascades were applied, incorporating LightGBM or XGBoost to enhance prediction accuracy. This combined approach offers a strong framework for the early detection of guava diseases and enhances classification precision.

\section{ Literature Review}

Guava (Psidium guajava) production is seriously threatened by diseases like fruit flies and anthracnose, which makes accurate classification techniques essential when utilizing the (GFDD24) dataset.  SqueezeNet was used by Kilci and Koklu (2024) \cite{r18} to extract deep features, and then AdaBoost and Gradient Boosting were used for classification; the accuracy of Gradient Boosting was 95.6\%.  Their results show that an efficient method for the early diagnosis of guava illnesses is to combine deep learning with ensemble machine learning techniques.

Clustering data with complicated patterns, including multi-modal or asymmetric distributions, is still a major difficulty in unsupervised learning using the GFDD24 dataset.  In order to solve this, Hong et al. (2025) \cite{r19} developed the Flexible Dirichlet Mixture Model (FDMM), which uses the technique of moments in conjunction with the EM algorithm for parameter estimation.  Multiple datasets were used to empirically confirm FDMM's improved flexibility and clustering performance, especially for complicated data distributions.

In the opinion of author Mostafa et al. (2021) \cite{r1} that guava disease detection methods highlighted Pakistan's significant role as one of the leading guava producers globally. The researchers proposed a novel approach employing deep learning (DL) techniques to automatically identify guava diseases. By employing color histograms and an unsharp filtering method, the data underwent a thorough cleaning process. Subsequently, they expanded and enhanced the dataset using affine transformations across nine angles with five deep learning networks were implemented in their study. Remarkably, ResNet-101 emerged as the most accurate model, achieving an impressive 97.74\% success rate in classifying guava diseases.

In a different context, Nandi (2022) \cite{r3} explored the efficiency of five renowned picture recognition models—VGG-16, GoogleNet, Resnet-18, MobileNet-v2, and EfficientNet. Among these, EfficientNet, with a size of 4.2MB and an accuracy of 99\%, outperformed the others, emphasizing its superior performance in terms of both size and accuracy. A comprehensive study \cite{r4} conducted in Bangladesh involved gathering images of guavas affected by anthracnose, fruit rot, and fruit canker. Three CNN models were employed, with the third model exhibiting exceptional performance, boasting an accuracy of 95.61\%.

Further investigations of Mumtaz et al. \cite{r5} utilized Darknet-53, AlexNet, and SidNet for feature extraction, followed by classification using SVM and KNN algorithms. The Quadratic SVM classifier emerged as the most successful, achieving an impressive accuracy of 98.9\% based on the selected features. In another intriguing study \cite{r6}, advanced models such as Fine KNN, Cubic SVM, Complex tree, boosted tree, and Bagged tree ensemble were employed for image-level and disease-level classification. The Bagged tree ensemble classifier stood out, delivering a remarkable overall accuracy of 99\%.

The utilization of deep-learning models DOUTOUM et al (2023) \cite{r7} VGG-16, Inception V3, ResNet50, and EfficientNet-B3 for identifying guava conditions showcased EfficientNet-B3 as the most accurate, with a training set accuracy of 97.83\%, a testing set accuracy of 92.73\%, and a test set accuracy of 94.93\%. Shifting gears, another study Perumal et al. (2021) \cite{r8} employed the K-Means Algorithm for segmentation, Gray Level Co-Occurrence Matrix (GLCM) for feature extraction, and SVM for classification, achieving an accuracy of 98.17\%. Yet another research endeavor Susai  et al.  (2023) \cite{r9} incorporated K-Means Algorithm for color segmentation and various feature extraction methods, including Shape Detection, GLCM, Grey Level Histogram, and Color Co-occurrence Matrix (CCM). This method yielded a commendable 96.67\% accuracy in assessing disease seriousness.

Szegedy et al. (2013) \cite{r14}, explored deep neural networks using a DNN regression approach capable of generating binary object copies and learning geometric features, validated on 5,000 images from the Pascal VOC 2007 dataset. Similarly, Habib et al. (2022) \cite{r15}, investigated jackfruit pest recognition using 480 images segmented with k-means and classified with Logistic Regression, SVM, Random Tree, RIPPER, KNN, Naive Bayes, BPN, and CPN. Their results showed Random Forest achieving the highest accuracy of 92.5\%, while KNN recorded the lowest at 70.42\%. These studies emphasize the effectiveness of both deep learning and traditional machine learning in agricultural image classification.

\section{Datasets Analysis}

\begin{figure}[!h]
    \centering
    \includegraphics[width=0.68\columnwidth]{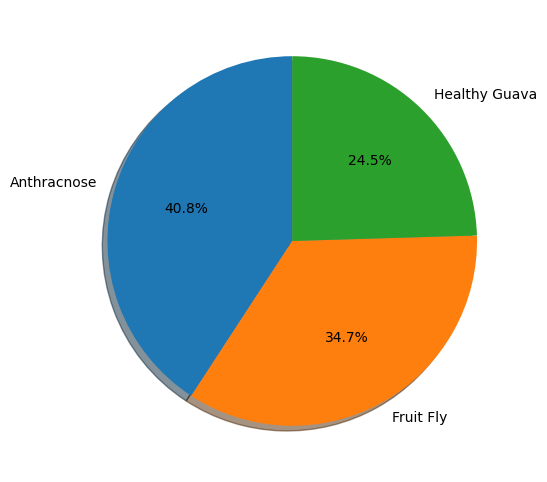}
    \caption{Guava Image Distribution}
    \label{fig:guava_image_distribution}
\end{figure}

The 3,784 photos in the GFDD24 collection are divided into three categories: Fruit Flies, Anthracnose, and Healthy Fruits shown in figure~\ref{fig:guava_image_distribution}. The dataset's producers carried out all preprocessing and augmentation procedures that increased its size from 473 raw photos to 3,784 samples before it was made public.  Originally released in distinct train, validation, and test folders, the dataset was combined into a single dataset for uniform testing.  To avoid any kind of data leaking, a new split was created from this combined dataset, allocating 80\% of the data for training and 20\% for validation. The original test set was kept only for the final assessment.

Class imbalance was discovered after the merging process, with certain classes having less samples than others, according to a study of the class distribution.  We found the smallest class size and used it as the reference count in order to remove class-based bias in model training.  To ensure equal representation for all three categories during training, classes with greater sample sizes were then randomly undersampled to match this lowest class.  This balancing strategy increased the consistency of the model's performance across classes and enabled each class to contribute equally to the learning process.

\begin{figure*}[t]
    \centering
    \begin{subfigure}[b]{0.32\textwidth}
        \centering
        \includegraphics[width=0.8\linewidth]{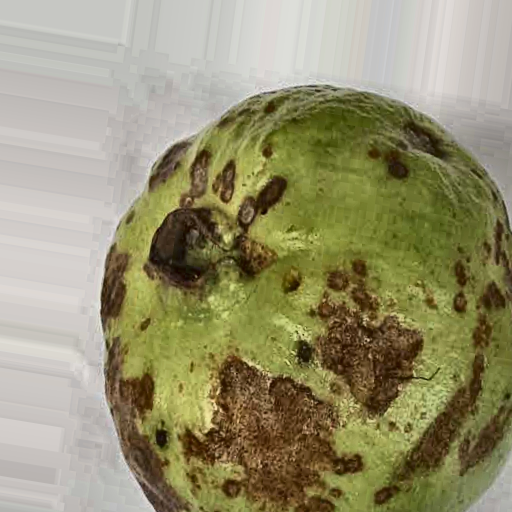}
        \caption{Anthracnose}
        \label{fig:Anthracnose}
    \end{subfigure}
    \hfill
    \begin{subfigure}[b]{0.32\textwidth}
        \centering
        \includegraphics[width=0.8\linewidth]{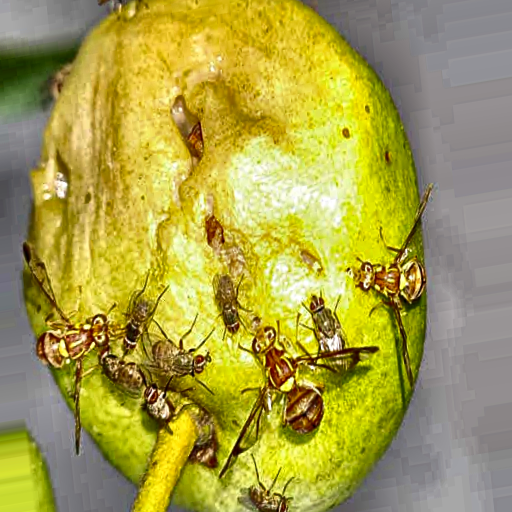}
        \caption{Fruit Fly}
        \label{fig:fruit_fly}
    \end{subfigure}
    \hfill
    \begin{subfigure}[b]{0.32\textwidth}
        \centering
        \includegraphics[width=0.8\linewidth]{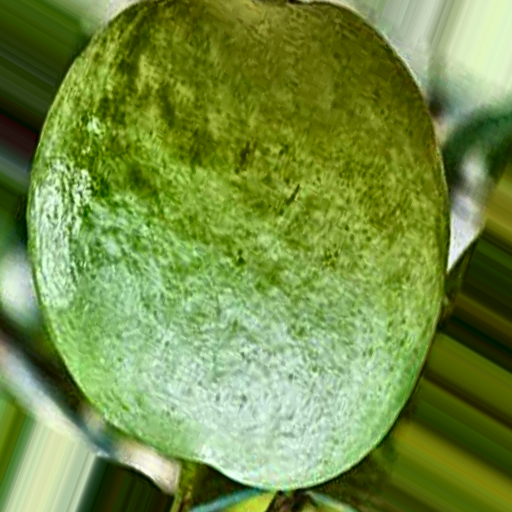}
        \caption{Healthy Guava}
        \label{fig:healthy_guava}
    \end{subfigure}
    \caption{Examples of guava conditions: (a) Anthracnose, (b) Fruit Fly, and (c) Healthy Guava.}
    \label{fig:guava_conditions}
\end{figure*}

\section{Dataset Preprocessing and Normalization}

Before making the dataset publicly available, the dataset authors finished all preprocessing operations related to GFDD24, such as picture downsizing to 512 × 512 pixels, RGB normalization, CLAHE, unsharp masking, and augmentation.  No further preprocessing, normalization, or augmentation techniques were used in our methodology.  The photos were utilized precisely as supplied, preserving complete compliance with the original dataset structure, except from combining the dataset and creating the new 80–20 training–validation split.

We highlight that, in contrast to our study, which only used random undersampling to resolve class imbalance, augmentation was carried out exclusively by the dataset producers prior to distribution. There was no oversampling or creation of new images.  The apparent inconsistency is resolved by this distinction: undersampling was used subsequently to obtain balanced class representation, whereas augmentation extended the dataset before our tests.  The experimental setting guarantees data integrity and removes the risk of leakage between training and assessment phases because the original test set was not altered during the whole process. As seen in figure~\ref{fig:guava_conditions}, these procedures provide a solid foundation for reliable model training and effective feature extraction using both deep learning and traditional machine learning techniques.

\subsection{Image Preprocessing and Generation}

As part of our preprocessing approach, we scaled each image to 224 × 224 and obtained a batch size of 32. Images from the dataset directory were loaded using an ImageDataGenerator with pixel rescaling (rescale=1./255).

\subsection{Features Extraction}
Due to its great representational power and efficiency, EfficientNet-B0 was selected as the CNN backbone for feature extraction.  Pretrained ImageNet weights are used to take advantage of transfer learning, and the top layers are eliminated to enable bespoke classification appropriate for the Guava Fruit Disease dataset.  A Global Average Pooling layer summarizes the data in a condensed representation while maintaining key characteristics by reducing the spatial dimensions following the creation of the convolutional feature maps.  Class probabilities are produced by a thick softmax layer, which is the last classification layer.  For multi-class classification problems, the model is constructed using the Adam optimizer with categorical crossentropy loss.

Let the input image be denoted as
\[
x \in \mathbb{R}^{H \times W \times C},
\]
where $H$, $W$, and $C$ are the height, width, and number of channels (RGB = 3) of the guava image.  

EfficientNet-B0 applies a series of convolutional operations $f(\cdot)$ parameterized by weights $W_c$ to extract deep feature maps:
\[
F = f(x; W_c), \quad F \in \mathbb{R}^{h \times w \times d},
\]
where $h$, $w$ are the reduced spatial dimensions after convolution and pooling, and $d$ is the number of filters (feature channels).  

Instead of flattening, a Global Average Pooling (GAP) layer reduces each feature map into a single representative value:
\[
z_j = \frac{1}{h \cdot w} \sum_{p=1}^{h} \sum_{q=1}^{w} F_{p,q,j}, 
\quad j = 1, 2, \ldots, d.
\]

This produces a condensed feature vector:
\[
z \in \mathbb{R}^d.
\]

The feature vector $z$ is then passed through a fully connected layer with weights 
$W_f \in \mathbb{R}^{d \times K}$ and bias $b_f \in \mathbb{R}^K$, where $K$ is the number of disease classes:
\[
u = W_f^T z + b_f, \quad u \in \mathbb{R}^K.
\]

The probability of class $k$ is computed using the softmax activation:
\[
P(y = k \mid x) = \frac{\exp(u_k)}{\sum_{j=1}^{K} \exp(u_j)},
\]
which ensures
\[
\sum_{k=1}^K P(y = k \mid x) = 1.
\]

For training, the categorical cross-entropy loss is minimized:
\[
L = -\frac{1}{N} \sum_{i=1}^N \sum_{k=1}^K y_{i,k} 
\log \big( P(y = k \mid x_i) \big),
\]
where $N$ is the number of training samples and $y_{i,k}$ is the one-hot encoded ground truth label.  

The Adam optimizer is used to iteratively update the model parameters by minimizing $L$:
\[
\theta_{t+1} = \theta_t - \eta \cdot 
\frac{\hat{m}_t}{\sqrt{\hat{v}_t} + \epsilon},
\]
where $\eta$ is the learning rate, and $\hat{m}_t$, $\hat{v}_t$ are the bias-corrected estimates of the first and second moments of the gradients.

\subsection{Data Balancing}

The GFDD24 dataset is imbalanced and we counted the smallest number of images in all classes by using Random Undersampling, to ensure a balanced representation. The sampled images and their annotations were combined to create a balanced dataset. To analyze the effects of sampling approaches, we examined our model using both balanced and unbalanced variants of the dataset.

\section{METHODOLOGY}

\begin{figure}[H]
    \centering
    \includegraphics[width=0.9\linewidth]{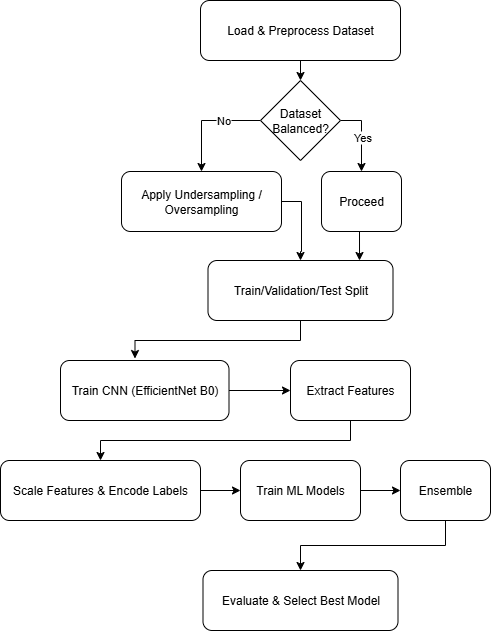}
    \caption{Working process of Dataset}
    \label{fig:dataset_process}
    
\end{figure}

\section{MODEL ANALYSIS }

\subsection{Proposed Ensemble Learning Models}

We implemented a hybrid ensemble learning framework to improve the accuracy and robustness of guava disease classification.  In a two-step procedure, the method integrates many classical and gradient boosting models.  Initially, the complete dataset is used to train a main classifier, such as Random Forest or AdaBoost.  We use class weighting to rectify class imbalances and ensure that every illness category makes an equivalent contribution during training. Predictions are assessed for confidence during inference, and samples with low confidence (less than 0.8) are regarded as uncertain. This hierarchical ensemble guarantees strong performance across all illness categories, minimizes misclassifications, and capitalizes on each model's unique capabilities.  Accuracy, weighted F1-score, precision, recall, and training and inference periods were collected to assess computational efficiency and performance.

Let the training dataset be defined as:
\[
D = \{(x_i, y_i)\}_{i=1}^N, \quad x_i \in \mathbb{R}^d, \; y_i \in \{1,2,\dots,C\},
\]
where $x_i$ is the $d$-dimensional feature vector, $y_i$ is the true class label, $N$ is the number of samples, and $C$ is the number of guava disease categories.

We first train a base classifier $f_{\text{base}}$ (e.g., Logistic Regression, Naive Bayes, KNN, SVM, Decision Tree, Random Forest, or AdaBoost) on the dataset $D$. The classifier outputs a probability distribution over all classes:
\[
p_{\text{base}}(y \mid x) = f_{\text{base}}(x), \qquad p_{\text{base}}(y \mid x) \in [0,1],
\]
\[
\sum_{c=1}^C p_{\text{base}}(y=c \mid x) = 1.
\]

The predicted class is chosen as:
\[
\hat{y}_{\text{base}} = \arg\max_{c \in \{1,\dots,C\}} p_{\text{base}}(y=c \mid x).
\]

To quantify prediction confidence, we define:
\[
\gamma(x) = \max_{c} p_{\text{base}}(y=c \mid x).
\]

If $\gamma(x) \geq \tau$, where $\tau = 0.8$ is the confidence threshold, the prediction from the base classifier is accepted:
\[
\hat{y}(x) = \hat{y}_{\text{base}}, \quad \text{if } \gamma(x) \geq \tau.
\]

For uncertain samples where $\gamma(x) < \tau$, the instance is passed to a CatBoost classifier $f_{\text{cat}}$, which outputs:
\[
p_{\text{cat}}(y \mid x) = f_{\text{cat}}(x), \qquad 
\hat{y}_{\text{cat}} = \arg\max_{c \in \{1,\dots,C\}} p_{\text{cat}}(y=c \mid x).
\]

Thus, the final hybrid ensemble decision rule is:
\[
\hat{y}(x) =
\begin{cases}
\hat{y}_{\text{base}}, & \text{if } \gamma(x) \geq \tau, \\[6pt]
\hat{y}_{\text{cat}}, & \text{if } \gamma(x) < \tau.
\end{cases}
\]

The objective of this hybrid system is to minimize the empirical misclassification risk:
\[
R(\hat{y}) = \frac{1}{N} \sum_{i=1}^N \mathbf{1}\big(\hat{y}(x_i) \neq y_i\big),
\]
while maintaining computational efficiency. The ensemble is evaluated in terms of accuracy, weighted F1-score, precision, recall, training time, and inference time.

\section{EXPERIMENTAL ENVIRONMENT}

This research was carried out on Google Colab using an NVIDIA T4 GPU, which provided ample computational power for CNN-based deep feature extraction, machine learning model training, and ensemble model evaluation. The cloud environment ensured consistent execution of the entire pipeline, enabling stable performance and reproducibility across all testing stages. 

A batch size of 32, 30 training epochs, the Adam optimizer, an image resolution of 224×224, and a validation split of 0.2 were manually configured as key hyperparameters. For feature extraction, the EfficientNet-B0 backbone was frozen, and a confidence threshold of 0.8 was applied within the hybrid ensemble framework. Most classical and boosting algorithms—including Logistic Regression, SVM, Random Forest, XGBoost, LightGBM, and CatBoost—were executed using their default configurations with minimal adjustments to ensure convergence and reproducibility. As a result, the reported model performance largely reflects baseline parameter settings, indicating significant room for enhancement through systematic hyperparameter tuning and more advanced optimization strategies in future work.

\section{Model Performance Analysis }

To thoroughly analyze the predictive quality and computational efficiency of the hybrid ensemble models, the following metrics were used: accuracy, weighted F1-score, precision, recall, training time, and inference time.

The AdaBoost-based hybrid models (Ada-LGBM, Ada-XGB, and Ada-Cat) obtained nearly flawless classification results on all assessment metrics, with $\sim$99.99\% accuracy, F1-score, precision, and recall, without class balancing shown in Table~\ref{tab:hybrid_models_without_balancing}. Random Forest-based hybrids, on the other hand, performed somewhat worse, averaging between 95-96\% percent accuracy with steady precision-recall tradeoffs. Despite its great predictive performance, Ada-Cat and RF-Cat's training times were noticeably longer than those of boosting-based combinations, indicating computational inefficiency.

A similar pattern continued when class balancing was used, with AdaBoost-driven hybrids once more producing nearly flawless predictions $\sim$99.99\% accuracy across all measures, demonstrating the consistency of their performance in balanced settings shown in Table~\ref{tab:hybrid_models_with_balancing}.  The benefits of balancing were demonstrated by Random Forest-based models, which showed slight increases in accuracy and recall when compared to the unbalanced situation.  However, even with their competitive prediction accuracy, CatBoost enhancements still imposed substantial training overheads.

CatBoost refinements imposed very high computational overheads, requiring \SI{582.33}{\second} of training and \SI{0.1494}{\second} of inference for RF-Cat and \SI{590.92}{\second} of training and \SI{0.2745}{\second} of inference for Ada-Cat on the imbalanced dataset in Table~\ref{tab:hybrid_models_without_balancing}, and \SI{877.39}{\second} of training and \SI{0.2623}{\second} of inference for RF-Cat and \SI{900.10}{\second} of training and \SI{0.3082}{\second} of inference for Ada-Cat on the balanced dataset in Table~\ref{tab:hybrid_models_with_balancing}, despite competitive accuracy (\numrange{95.23}{99.99}\,\%). 
XGBoost models offered intermediate performance, with RF-XGB achieving \numrange{96.23}{96.42}\,\% accuracy and moderate training/inference times, while Ada-XGB reached \num{\sim 0.9999} (\textasciitilde99.99\,\%) accuracy at higher training costs. 
RF-LGBM provided the best trade-off, achieving \numrange{96.23}{96.42}\,\% accuracy with only \textasciitilde\SI{35.8}{\second} of training and \SIrange{0.053}{0.061}{\second} of inference, making it the most practical choice for real-time or resource-constrained applications.

AdaBoost-based hybrids, especially those in conjunction with LightGBM or XGBoost, are the most dependable choices if accuracy is the main concern. They often achieve $\sim$99.99\% across all measures.  However, Random Forest-based hybrids with LightGBM or XGBoost offer a better fair trade-off if computational economy is the key concern. They retain good performance while taking a lot less time for training and inference than CatBoost-based improvements.

\begin{table}[H]
    \centering
    \renewcommand{\arraystretch}{1.6}
    \caption{Performance Metrics of Hybrid Models Without Class Balancing}
    \resizebox{\columnwidth}{!}{ 
        \begin{tabular}{|l|c|c|c|c|c|c|}
            \hline
            \textbf{Model}   & \textbf{Accuracy} & \textbf{F1-score} & \textbf{Precision} & \textbf{Recall} & \textbf{Training (s)} & \textbf{Inference (s)} \\ \hline
            RF-LGBM          & 0.9623            & 0.9519            & 0.9541             & 0.9523          & 35.8412               & 0.0527                \\ \hline
            Ada-LGBM         & 0.9999            & 0.9999            & 0.9999             & 0.9999          & 137.8054              & 0.2046                \\ \hline
            RF-XGB           & 0.9623            & 0.9519            & 0.9541             & 0.9523          & 41.8793               & 0.0318                \\ \hline
            Ada-XGB          & 0.9999            & 0.9999            & 0.9999             & 0.9999          & 107.8885              & 0.1814                \\ \hline
            RF-Cat           & 0.9523            & 0.9519            & 0.9541             & 0.9523          & 582.3289              & 0.1494                \\ \hline
            Ada-Cat          & 0.9999            & 0.9999            & 0.9999             & 0.9999          & 590.9168              & 0.2745                \\ \hline
        \end{tabular}
    }
    \label{tab:hybrid_models_without_balancing}
\end{table}

\begin{table}[H]
    \centering
    \renewcommand{\arraystretch}{1.6}
    \caption{Performance Metrics of Hybrid Models With Class Balancing}
    \resizebox{\columnwidth}{!}{ 
        \begin{tabular}{|l|c|c|c|c|c|c|}
            \hline
            \textbf{Model}   & \textbf{Accuracy} & \textbf{F1-score} & \textbf{Precision} & \textbf{Recall} & \textbf{Training (s)} & \textbf{Inference (s)} \\ \hline
            RF-LGBM          & 0.9642            & 0.9394            & 0.9482             & 0.9642          & 35.8739               & 0.0607                \\ \hline
            Ada-LGBM         & 0.9999            & 0.9999            & 0.9999             & 0.9999          & 113.4373              & 0.1565                \\ \hline
            RF-XGB           & 0.9642            & 0.9394            & 0.9482             & 0.9642          & 37.4522               & 0.0423                \\ \hline
            Ada-XGB          & 0.9999            & 0.9999            & 0.9999             & 0.9999          & 90.5174               & 0.1919                \\ \hline
            RF-Cat           & 0.9534            & 0.9534            & 0.9551             & 0.9534          & 877.3886              & 0.2623                \\ \hline
            Ada-Cat          & 0.9999            & 0.9999            & 0.9999             & 0.9999          & 900.0965              & 0.3082                \\ \hline
        \end{tabular}
    }
    \label{tab:hybrid_models_with_balancing}
\end{table}

\begin{figure}[H]
    \centering
    \includegraphics[width=\columnwidth]{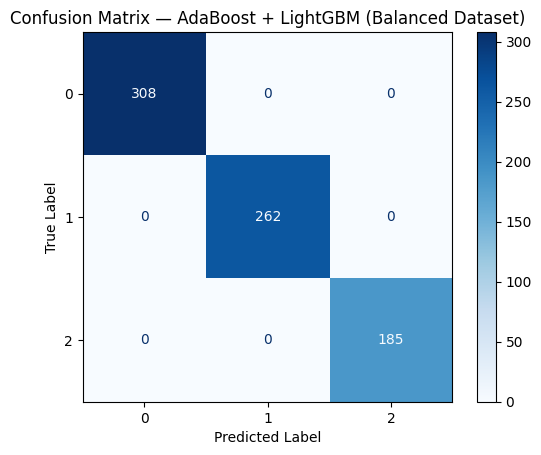}
    \caption{Confusion Matrix}
    \label{fig:Confusion Matrix}
\end{figure}

To guarantee objective assessment across all classes, the AdaBoost-LightGBM hybrid model's confusion matrix is created using a balanced dataset shown in figure~\ref{fig:Confusion Matrix}. Because every class participates equally, this method ensures meaningful assessment that reflects the model's actual efficacy.  On this balanced dataset, the Ada-LGBM hybrid model shows 99.99\% accuracy, indicating dependable categorization across various disease categories.  As a result, the performance of the suggested system is accurately represented by the balanced confusion matrix.

Overall, the findings show that hybrid ensembles based on AdaBoost often perform better in classification reliability than other ensembles, especially when combined with gradient boosting models.  However, the trade-off between computing cost and accuracy should be taken into account for practical implementation, particularly for real-time or large-scale applications.

\section{Conclusion}

Leveraging the GFDD24 dataset, this study demonstrates the effectiveness of integrating convolutional neural networks with ensemble machine learning techniques for guava disease detection. Using EfficientNet-B0 for deep feature extraction with hybrid models such as LightGBM, XGBoost, and CatBoost, we achieved high accuracy in classifying anthracnose, fruit flies, and healthy fruits. Random Forest--based models also performed competitively with lower computational demands, while AdaBoost ensembles consistently reached $\sim$99.99\% accuracy, precision, recall, and F1-score. These results highlight the robustness of the proposed CNN--ML cascade framework for both balanced and unbalanced datasets. Overall, the approach provides a solid foundation for real-time guava disease monitoring systems, successfully integrating trained CNN models into a responsive web based detection platform. So, The strong performance suggests its suitability for real-world agricultural applications, enabling farmers to make timely decisions to reduce crop losses.

Although the current system effectively employs pure CNN models on responsive web and mobile platforms for detecting guava diseases, several improvements remain possible. The GFDD24 dataset lacks broad spatial and environmental variation, so future work will focus on expanding it across more seasons and regions. Upcoming research will also compare the performance of CNNs with transformer-based approaches such as ViT and the Swin Transformer. Moreover, efficiency on low-resource devices will be enhanced through optimization strategies like pruning and quantization. These advancements aim to build a more robust and scalable detection solution.

\bibliographystyle{IEEEtran}
\bibliography{references}
\end{document}